\documentclass{article}

\usepackage[final]{neurips_robustbayes21}

\usepackage{wrapfig}
\usepackage[utf8]{inputenc} % allow utf-8 input
\usepackage[T1]{fontenc}    % use 8-bit T1 fonts
\usepackage{hyperref}       % hyperlinks
\usepackage{url}            % simple URL typesetting
\usepackage{booktabs}       % professional-quality tables
\usepackage{amsfonts}       % blackboard math symbols
\usepackage{nicefrac}       % compact symbols for 1/2, etc.
\usepackage{microtype}      % microtypography
\usepackage{xcolor}         % colors
\usepackage{graphicx}
\usepackage{amsmath, amssymb}
\usepackage{amsfonts}
\usepackage{verbatim}

\usepackage{subfiles} % Best loaded last in the preamble

\usepackage{authblk}

\title{Uncertainty estimation under model misspecification in neural network regression}

\author[1]{\textbf{Maria R. Cervera*}}
\author[1]{\textbf{ Rafael Dätwyler*}}
\author[1]{\textbf{Francesco D'Angelo*}}
\author[1,2]{\textbf{Hamza Keurti*}}
\author[1]{\textbf{Benjamin F.~Grewe}}
\author[1,3]{\textbf{Christian Henning}}

\affil[ ]{*Equal Contribution}
\affil[1]{\footnotesize Institute of Neuroinformatics\\
  University of Zürich and ETH Zürich\\
  Zürich, Switzerland}
\affil[2]{\footnotesize Max Planck ETH Center for Learning Systems} %Tübingen, Germany
\affil[3]{\footnotesize Institute of Theoretical Computer Science, ETH Zürich, Zürich, Switzerland }
\affil[ ]{Correspondence: \texttt{\{mariacer,henningc\}@ethz.ch} }

\begin{document}

\maketitle
\vspace{-2em}
\begin{abstract}
\vspace{-1em}
Although neural networks are powerful function approximators, the underlying modelling assumptions ultimately define the likelihood and thus the hypothesis class they are parameterizing. In classification, these assumptions are minimal as the commonly employed softmax is capable of representing any categorical distribution. In regression, however, restrictive assumptions on the type of continuous distribution to be realized are typically placed, like the dominant choice of training via mean-squared error and its underlying Gaussianity assumption. Recently, modelling advances allow to be agnostic to the type of continuous distribution to be modelled, granting regression the flexibility of classification models. While past studies stress the benefit of such flexible regression models in terms of performance, here we study the effect of the model choice on uncertainty estimation. We highlight that under model misspecification, aleatoric uncertainty is not properly captured, and that a Bayesian treatment of a misspecified model leads to unreliable epistemic uncertainty estimates. Overall, our study provides an overview on how modelling choices in regression may influence uncertainty estimation and thus any downstream decision making process.
\end{abstract}

\vspace{-2em}
\section{Introduction}
\label{sec:intro}

\vspace{-.5em}
We revisit the problem of supervised learning with neural networks via a dataset $\mathcal{D} \sim p(X) p(Y \mid X)$ where outcomes $Y \in \mathcal{Y} \subseteq \mathbb{R}^{d_\text{out}}$ are continuous. The application of neural networks to such regression problems requires making modelling assumptions on the type of likelihood function being considered, and these have profound implications on uncertainty estimation and downstream decision making. In this study we review and demonstrate the consequences of common assumptions and outline how modern modelling tools like normalizing flows \citep[NF, ][]{papamakarios2019nf:review} can mitigate past limitations regarding uncertainty estimation. As such, this study may serve as a compact set of considerations to be taken into account before applying deep learning tools to regression problems.

\begin{wrapfigure}{r}{0.49\textwidth}
  \begin{center}
  \vspace{-2mm}
    \includegraphics[width=0.47\textwidth]{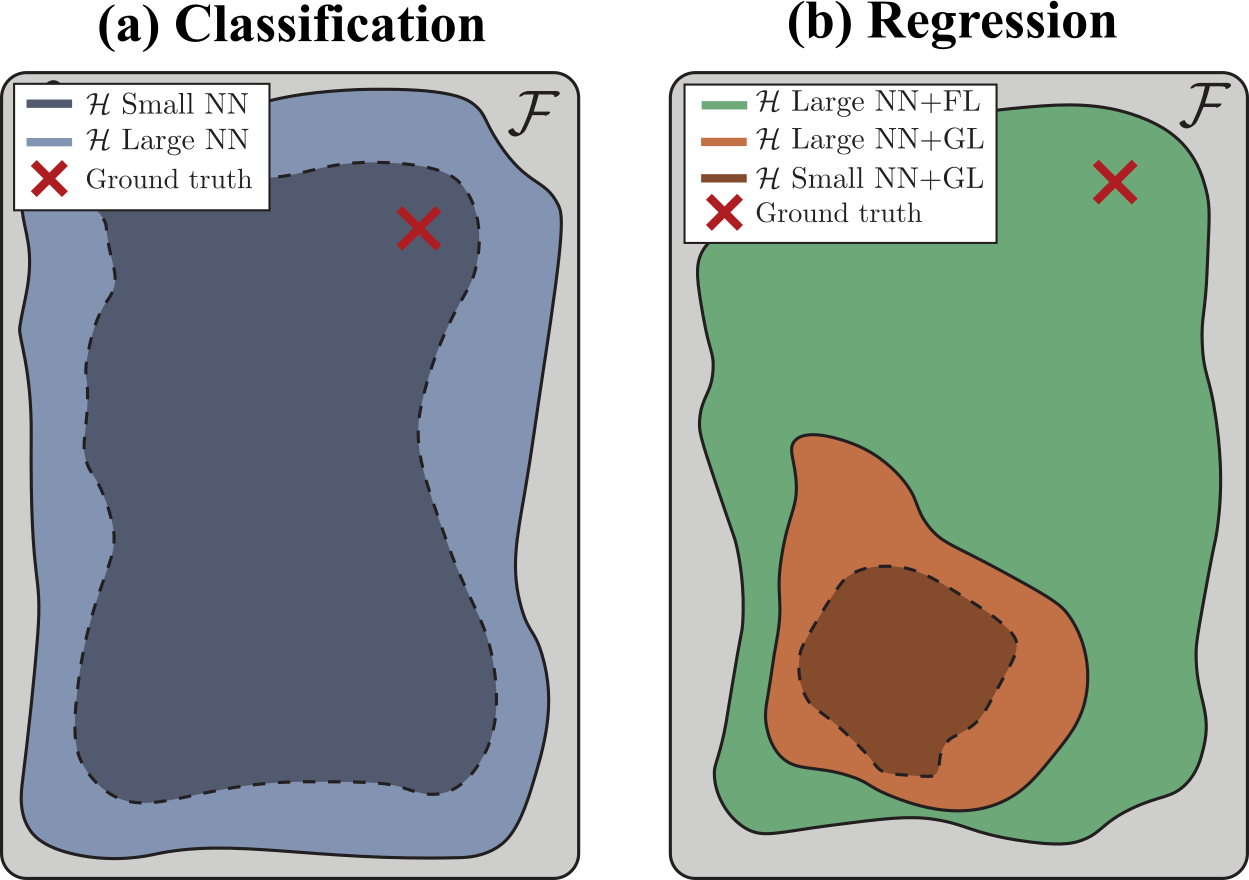}
  \end{center}
  \vspace{-2mm}
  \caption{
  Space of conditional distributions $p(Y \mid X)$. \textbf{(a)} If $\mathcal{Y} = \{1,\dots,C\}$, and the underlying neural network (NN) approaches a universal function approximator, any $p(Y \mid X)$ can be arbitrarily well approximated by a softmax output. \textbf{(b)} If $\mathcal{Y} \in \mathbb{R}^{d_\text{out}}$, common model assumptions like the convenient choice of restricting $p(Y \mid W, X)$ to be Gaussian severely restrict the expressiveness of the neural network-based model and might have detrimental consequences on predictions if modelling assumptions are incorrect.
  }
  \vspace{-4mm}
    \label{fig:hypothesis:space}
\end{wrapfigure}

In order to approximate the conditional $p(Y \mid X)$ of the unknown underlying data distribution, a model $p(Y \mid W, X)$ with parameters $W$ can be obtained by minimizing the expected KL divergence between the ground-truth and the estimated conditional distribution: $\mathbb{E}_{p(X)} \big[ \text{KL} \big( p(Y \mid X) || p(Y \mid W, X) \big) \big] $. This optimization criterion amounts to minimizing the expected cross-entropy $- \mathbb{E}_{p(X, Y)} [\log p(Y \mid W, X)]$ and can be approximated using the negative log-likelihood (NLL) of the data  $\mathcal{D}$: $ - \sum_{(x, y) \sim \mathcal{D}} \log p(y \mid W, x)$ (cf. SM \ref{sm:sec:loss}). This approach has led to impressive results in classification tasks, where the softmax output allows modelling any input-dependent discrete distribution arbitrarily well. Indeed, if $\mathcal{F}$ denotes the space of all conditional densities (i.e., functions $\mathcal{X} \times \mathcal{Y} \rightarrow [0, \infty)$, normalized in $\mathcal{Y}$), and $\mathcal{H} \subseteq \mathcal{F}$ is the hypothesis space defined by the choice of model 
$p(Y \mid W, X)$ and parametrized by $W$, an expressive neural network with softmax output will arguably cause $\mathcal{F}$ to be largely covered by $\mathcal{H}$ (Fig. \ref{fig:hypothesis:space}a).
By contrast, in regression there is no immediate softmax equivalent and assumptions on the distributional form of $p(Y \mid X)$ are necessary. A common strategy is to assume Gaussianity in $p(Y \mid X)$ and therefore define a likelihood with a Gaussian model $p(Y \mid W, X)$ (\textbf{GL}).
% See work mentioning this already: https://ieeexplore.ieee.org/stamp/stamp.jsp?tp=&arnumber=118274
More specifically, the variance is commonly assumed to be input independent $\mathcal{N}\big(Y;  f(X; W), \sigma^2 I \big)$ (i.e., homoscedastic noise model, \textbf{GLc}), which results in the widely established mean-squared error (MSE) loss function (cf. SM \ref{sm:sec:loss}). Under such assumptions, only a small subset of $\mathcal{F}$ can be covered by $\mathcal{H}$ (Fig. \ref{fig:hypothesis:space}b), which is highly problematic if the model is misspecified, i.e., $\mathcal{H}$ does not contain the ground-truth. Notably, this limitation is independent of the expressiveness of the employed neural network, and highlights fundamental differences in solving classification and regression tasks with neural networks.

The issues of inference under model misspecification have been studied in \cite{white1982maximum,grunwald2007suboptimal,grunwald2017inconsistency,ramamoorthi2015posterior} and more recently the PAC-Bayes framework has been deployed to study the consequences of misspecification on generalization \citep{morningstar2020pac,pacmasegosa2020}. Even more, to overcome the limitations arising from restrictive model assumptions, more flexible conditional density estimation approaches have been recently developed. In particular NFs have been deployed for regression problems \citep{trippe2018conditional, rothfuss2019noise, zieba2020regflow, sick2020deep}, providing a continuous substitution for the softmax that requires minimal assumptions about the distributional form of $p(Y \mid X)$, and leading to performance improvements in a range of tasks. In these flow-based models (\textbf{FL}), the flow's parameters are input-dependent (cf. Fig. \ref{fig:nets:different:likelihood:class}) to allow conditional density estimation.

Crucially, correctly capturing $p(Y \mid X)$ is not only important for making predictions, but also for accurately estimating how certain we are about those predictions. Uncertainty is commonly split into aleatoric and epistemic uncertainty. \textit{Aleatoric uncertainty} refers to the uncertainty intrinsic to the data, and can only be fully captured if $p(Y \mid X)$ is contained in $\mathcal{H}$. \textit{Epistemic uncertainty}, on the other hand, captures the uncertainty about having correctly identified the ground-truth from the limited data seen. This uncertainty can be further divided into approximation and model uncertainty \citep{hullermeier2021aleatoric:epistemic}. \textit{Approximation uncertainty} describes the uncertainty about which $h \in \mathcal{H}$ is closest to $p(Y \mid X)$ and can, for instance, be captured by applying Bayesian statistics to the network weights $W$. In contrast \textit{model uncertainty} captures the belief that the ground-truth is contained in $\mathcal{H}$. This type of uncertainty is in practice often ignored as it is difficult to estimate in the case of neural networks. However, while the assumption that $p(Y \mid X)$ lies in $\mathcal{H}$ is reasonable for classification, in common regression scenarios ignoring model uncertainty and equating epistemic uncertainty with approximation uncertainty can create the false impression that $p(Y \mid X)$ is faithfully captured and misguide decision making (cf. Sec. \ref{sec:exp}).

Here we compare the modelling choices GLc, GL and FL for regression problems (Fig. \ref{fig:nets:different:likelihood:class}), and demonstrate the effects of model misspecification on aleatoric and epistemic uncertainty estimates. Being aware of these effects is crucial for designing regression models since wrongly modelled aleatoric uncertainty may cause erroneous predictions, which may be assigned with high confidence when using common surrogates of epistemic uncertainty (such as approximation uncertainty).

\vspace{-1em}
\section{Results}
\label{sec:exp}
\vspace{-.5em}

\begin{figure}[t]
    \includegraphics[width=\textwidth]{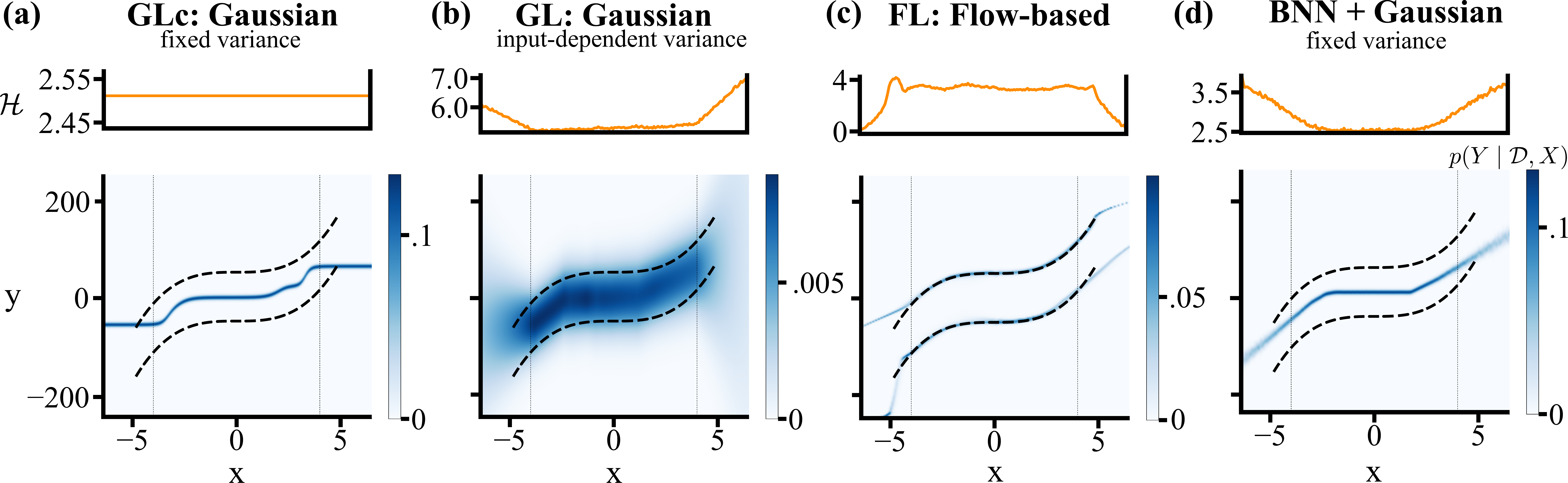}
    \vspace{-1em}
    \caption{Differential entropy $\mathcal{H}$ (upper row) and predictive distributions (lower row) for the 1D mixture of Gaussians experiments under different model classes. \textbf{(a)} Gaussian likelihood with fixed variance, \textbf{(b)} or varying variance, \textbf{(c)} normalizing flow-based likelihood, \textbf{(d)} Gaussian likelihood with fixed variance and parameter uncertainty. Dashed lines indicate the means of the two modes in the ground-truth, and the vertical dotted lines indicate the boundaries of the in-distribution range.}
    \label{fig:bimodal:1D:experiments}
    \vspace{-6mm}
\end{figure}

\begin{wrapfigure}{r}{0.35\textwidth}
  \begin{center}
  \vspace{-2em}
    \includegraphics[width=0.35\textwidth]{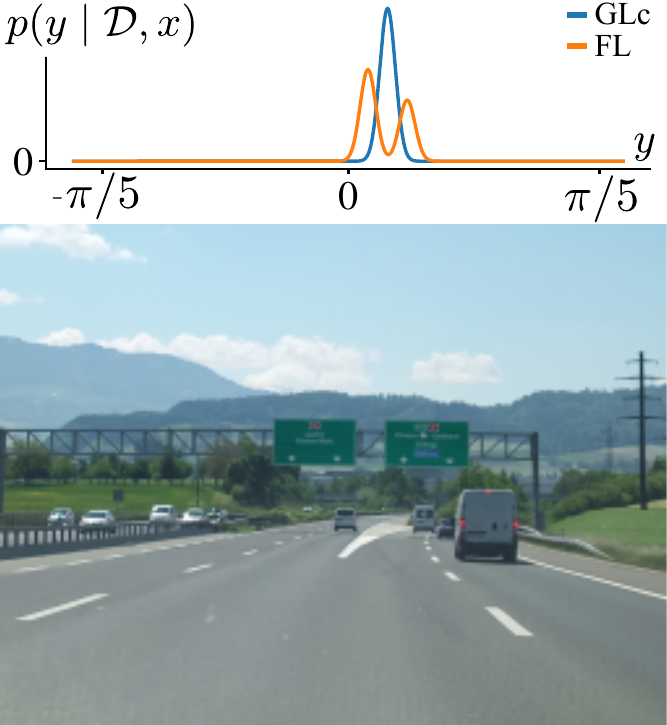}
  \end{center}
  \vspace{-2mm}
  \caption{When deployed to situations where several outcomes are possible, such as steering angle prediction when a lane splits in two, the FL model can capture multi-modality if such examples are sufficiently represented in the training data.
  By contrast, a GLc model would predict the mean of the two possible outcomes and crash the car in between lanes.}
  \vspace{-2mm}
    \label{fig:split}
\end{wrapfigure}

To visualize the behavior of different modelling choices, we first consider a 1D regression problem with known ground-truth $p(X, Y)$, where $p(Y \mid X)$ is a mixture of Gaussians (Fig. \ref{fig:bimodal:1D:experiments}, SM \ref{sm:exp:oned}). This problem represents a scenario with two possible outcomes (e.g., a car's steering command at a three-way junction, cf. Fig. \ref{fig:split}), for which the model classes GLc and GL are misspecified and cannot faithfully capture $p(Y \mid X)$. Given that any downstream algorithm for making predictions relies on the obtained estimate of $p(Y \mid X)$, it is of crucial importance to understand the effect that model misspecification has on the different types of uncertainty.

Aleatoric uncertainty can be quantified through a range of statistics on the ground-truth $p(Y\mid X)$, and can therefore only be properly captured if the model perfectly matches $p(Y \mid x)$ for $x \sim p(X)$. How well this is accomplished is commonly measured via calibration metrics \citep{guo2017calibration} using, for instance, the NLL on a test set $\mathcal{D}_{test}$ as a surrogate for measuring $\mathbb{E}_{p(X)} \big[ \text{KL} \big( p(Y \mid X) || p(Y \mid W, X) \big) \big]$. For GLc (Fig. \ref{fig:bimodal:1D:experiments}a), best calibration only allows a proper estimation of the ground-truth mean (cf. SM \ref{sm:sec:glc:aleatoric}), while statistics relevant for quantifying aleatoric uncertainty, such as the variance $\sigma^2$, are often chosen arbitrarily (as opposed to taking or estimating the true $\text{Var}_{p(Y \mid X)}\{Y\}$, cf. Fig. \ref{fig:bimodal:1D:true:std}). This can create the false impression of having high-confidence predictions, which in the steering angle prediction problem (Fig. \ref{fig:split}) would cause the car to confidently crash between two lanes. By contrast, best calibration of GL (Fig. \ref{fig:bimodal:1D:experiments}b) not only leads to an accurate estimation of the mean, but also of the variance of the ground-truth $p(Y \mid X)$. Although predictions from this model are still inaccurate (e.g., the car would still crash), the ground-truth variance can be properly captured. Note, however, that while high variance prevents confident predictions here, this statistic does not faithfully reflect the ground-truth's uncertainty, which can make confident predictions within each mode. Finally, when the model class contains the ground-truth (i.e., arguably FL) and $p(Y \mid x)$ is accurately captured for $x \sim p(X)$ by the model (Fig. \ref{fig:bimodal:1D:experiments}c), any in-distribution uncertainty quantification via $p(Y \mid W, X)$ (e.g., diff. entropy of the predictive distribution) can meaningfully represent aleatoric uncertainty. Moreover, samples from $p(Y \mid W, X)$ are useful predictions and ambiguities in those predictions can be detected by downstream decision making algorithms (e.g., a navigation system would be necessary to resolve the ambiguity stemming from a lane splitting situation).

Since the amount of available data $\mathcal{D}$ is limited, we cannot expect that the statistics of $p(Y \mid X)$ are faithfully captured by $p(Y \mid W, X)$ (cf. Fig. \ref{fig:bimodal:1D:20pts:experiments}). Uncertainty about how well $p(Y \mid X)$ is captured can be explicitly modelled by estimating epistemic uncertainty, which is often reduced in practice to estimating approximation uncertainty by treating the parameters $W$ as a random variable with prior distribution $p(W)$. The models which are in agreement with the data are summarized by the posterior parameter distribution $p(W \mid \mathcal{D})$, and predictions are made via marginalization, using the posterior predictive distribution $p(Y \mid \mathcal{D}, X) = \int_w p(w \mid \mathcal{D}) p(Y \mid w, X) \, dw$ (for an introduction see \citet{mackay2003itila}). This framework applied to neural networks leads to so-called Bayesian neural networks \citep[BNN, ][]{mackay:1992:practical}. Since Bayesian inference in this setting is intractable, approximate schemes such a variational inference \citep{bayes:by:backprop} are necessary. We explore the behavior of a BNN with a GLc likelihood in Fig.~\ref{fig:bimodal:1D:experiments}d (see SM \ref{sm:exp:oned} for more examples). While a detailed discussion on how to quantify approximation uncertainty is deferred to SM \ref{sm:sec:uncertainty}, it is apparent from the diff. entropy of the predictive distribution that our estimate of epistemic uncertainty has vanished in-distribution. This is to be expected, as the mean of $p(Y \mid X)$ has been correctly identified by the model. 
However, the application of Bayesian statistics should not give the false impression that it is safe to draw predictions via this model; the model class is misspecified and therefore model uncertainty, which is not taken into consideration, is nonzero. 

To highlight the limitations of MSE training for capturing correlations within $p(Y \mid X)$, we also consider a multivariate regression problem consisting of a bimodal data distribution $p(Y \mid X)$ in a 2D output space (Fig.~\ref{fig:bimodal:2D:experiments}). Assuming a GLc model class in a multivariate setting, leads to an independence assumption among dependent variables, which may cause inaccurate predictions and can be problematic in many real-world scenarios (e.g., control of robotic arms with multiple joints). While this problem can be mitigated by using a multivariate Gaussian with input-dependent covariance matrix (GL), full flexibility can be achieved with the FL model. Interestingly, our dataset is designed such that the axis-aligned projections appear like independent Gaussians (Fig.~\ref{fig:bimodal:2D:experiments}), highlighting the risks of making simplistic modelling choices after a superficial scan of the data.

\vspace{-1em}
\section{Discussion}
\vspace{-.5em}

Faithfully capturing $p(Y \mid X)$ within the support of $p(X)$ is crucial for obtaining reliable predictions with discriminative models. In this work, we outlined the particular challenges that arise when $Y$ is continuous, and highlight the importance of correctly interpreting uncertainty estimates when judging a model's reliability.

\textbf{Uncertainty estimation.}
Access to accurate predictions and to reliable aleatoric uncertainty estimates requires the model to match $p(Y \mid X)$ in-distribution, which is only possible if the hypothesis class $\mathcal{H}$ contains the ground-truth.
In practice, $\mathcal{H}$ is implicitly often reduced to a model with Gaussian likelihood without proper justification or validation. We discussed how this choice affects measures of aleatoric and epistemic uncertainty, and showed how modern modelling tools like normalizing flows can alleviate past shortcomings.

\textbf{Computational considerations.} Using a flow-based likelihood function has obvious computational drawbacks in terms of memory, runtime and optimization complexity. For instance, a more complex model class is arguably more data-hungry for identifying the ground-truth $p(Y \mid X)$. Therefore, modellers have to carefully trade-off expressiveness with computational benefits. 

\textbf{Independent and identically distributed samples.} Another source of misspecification to be kept in mind when designing a model is the ubiquitous underlying assumption regarding the availability of i.i.d. samples from $p(X) p(Y \mid X)$ (which, for instance, allows for a fully-factorized likelihood, cf. SM \ref{sm:sec:loss}). Note that, although we do not focus on this aspect here, this assumption is for example violated in the steering angle prediction problem we consider (SM \ref{sm:exp:udacity}), which contains only a short amount of recorded human driving experience.

\textbf{Out-of-distribution (OOD) detection.}
In contrast to softmax-based classifiers, GLc regression models do not exhibit input-dependent uncertainty, and are thus not applicable for uncertainty-based OOD detection (Fig. \ref{fig:bimodal:1D:experiments}a). However, in a Bayesian setting, even such simple models exhibit input-dependent uncertainty and can, in simple 1D regression tasks such as in Fig. \ref{fig:unimodal:1D:experiments}d, give the impression of being good OOD detection tools \citep[e.g., ][]{louizos:multiplicative:nf:2017}. In this context, it is important to keep in mind that OOD uncertainty is arbitrary since the NLL loss only calibrates uncertainties in-distribution, and even more flexible models like FL behave arbitrarily on OOD data (Fig. \ref{fig:bimodal:1D:experiments}c). For a general discussion on the limitations of BNNs for OOD detection please refer to \cite{henning2021bayesian}. We discuss this topic in the context of regression problems in SM \ref{sm:sec:ood}.

\textbf{Conclusion.} While modern deep learning tools may alleviate the requirement for strong modelling assumptions, the complexity associated with using these tools (e.g., the application of approximate Bayesian inference methods) may ultimately harm uncertainty estimation and predictive quality. For practitioners, it is important to keep all these considerations in mind and be aware of limitations such that a solution that captures the data well is found and the resulting model becomes as useful as it can be. We hope that our elaborations in this study help researchers and engineers identify the intricacies associated with regression problems, and ultimately lead to better modelling choices. 

\acksection
This work was supported by the Swiss National Science Foundation (B.F.G. CRSII5-173721 and 315230\_189251), ETH project funding (B.F.G. ETH-20 19-01) and funding from the Swiss Data Science Center (B.F.G, C17-18).
Hamza Keurti is supported by the Max Planck ETH Center for Learning Systems.

\bibliographystyle{plainnat}
\bibliography{bibliography}

%%%%%%%%%%%%%%%%%%%%%%%%%%%%%%
%%% Supplementary material %%%
%%%%%%%%%%%%%%%%%%%%%%%%%%%%%%
\newpage
\normalsize
\setcounter{page}{1}
\setcounter{figure}{0} \renewcommand{\thefigure}{S\arabic{figure}}
\setcounter{table}{0} \renewcommand{\thetable}{S\arabic{table}}
\appendix

\section*{\Large{Supplementary Material: Uncertainty estimation under model misspecification in neural network regression}}
\textbf{Maria R. Cervera*, Rafael Dätwyler*, Francesco D'Angelo*, Hamza Keurti*, Benjamin F.~Grewe, Christian Henning} \\

\section{Summary of modeling choices}

\begin{figure}[h]
    \centering
    \includegraphics[width=.85\textwidth]{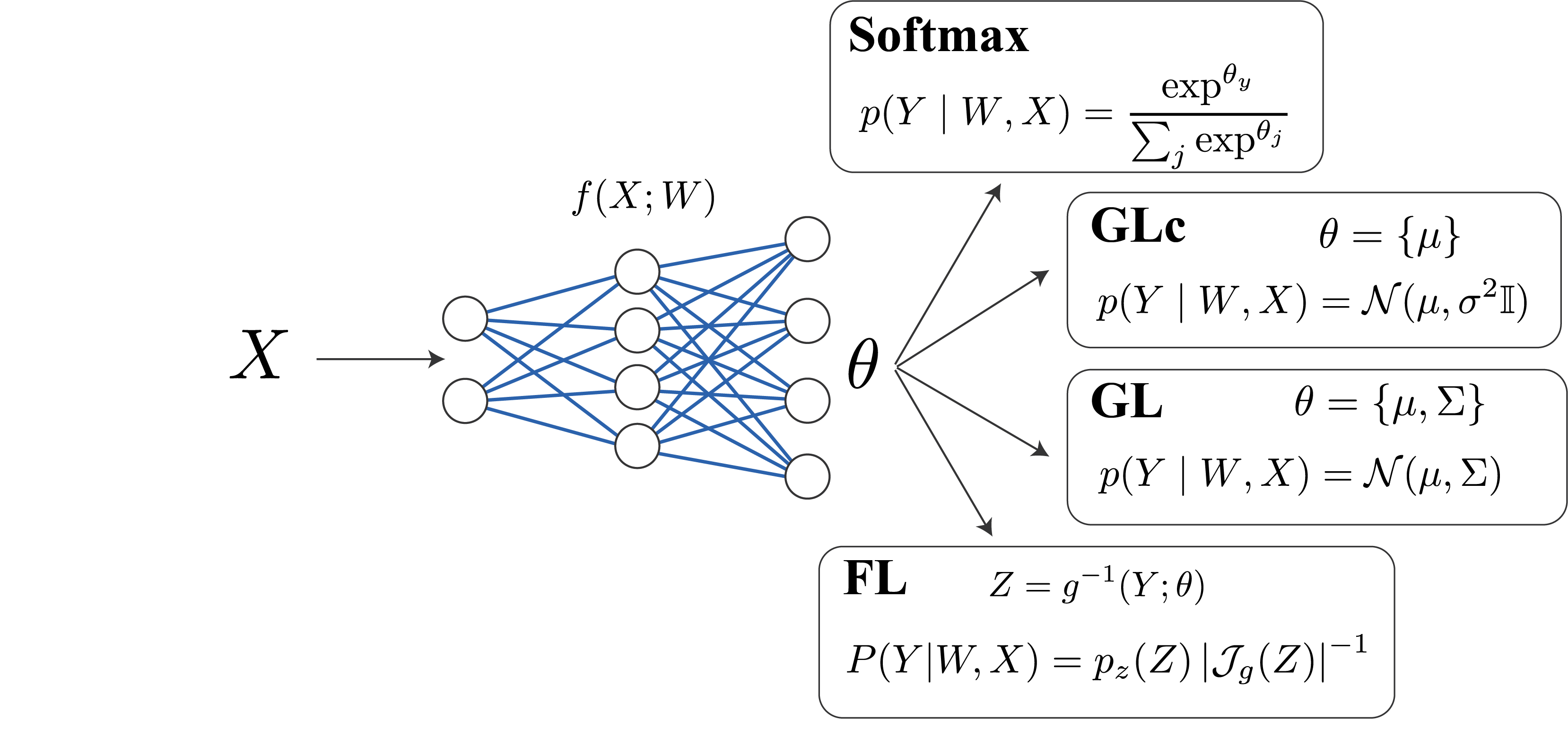}
    \caption{This figure illustrates how neural networks can be used to produce different hypothesis spaces. Intuitively, the neural network outputs the $x$-dependent parameters $\theta = f(X; W)$ of a conditional density $p(Y \mid W, X) \equiv p\left(Y \mid f(X; W)\right)$. In classification, $\theta$ is commonly processed by a softmax to produce a categorical distribution over class labels. In most regression tasks, $\theta$ represents the mean $\mu$ of a Gaussian distribution (GLc). If $\theta$ encodes both mean $\mu$ and covariance matrix $\Sigma$, then the model can exhibit input-dependent uncertainty (GL). To overcome strong assumptions on the distributional form of $p(Y \mid X)$ in the continuous case, flow-based likelihood functions (FL) have recently been proposed.}
    \label{fig:nets:different:likelihood:class}
\end{figure}

We summarize and compare the different modeling choices considered in this study in Fig. \ref{fig:nets:different:likelihood:class}.

\section{Loss functions}
\label{sm:sec:loss}
Regression aims to find the model for which the conditional density $p(Y|W,X)$ matches best the true conditional density $p(Y|X)$. This is commonly achieved by minimizing the expected forward KL divergence between $p(Y|X)$ and $p(Y|W,X)$: 
\begin{align}
    & \min_W \quad E_{p(X)}\big[KL(p(Y|X)||p(Y|W,X))\big]\\
    % \Leftrightarrow & \min_W \quad E_{p(Y,X)}\big[E_{p(X)}\log\big(\frac{p(Y|X)}{p(Y|W,X)}\big)\big].\\
    % \Leftrightarrow & \min_W\quad E_{p(X)}\big[E_{p(Y|X)}\log p(Y|X) - E_{p(Y|X)}\log p(Y|W,X)\big].\\
    \Leftrightarrow & \min_W \quad  -E_{p(X)}\big[H(p(Y|X))\big] - E_{p(X,Y)}\big[\log p(Y|W,X)\big]\label{eq:entropyin}\\
    \Leftrightarrow & \min_W \quad  -E_{p(X,Y)}\big[\log p(Y|W,X)\big]
    % \Leftrightarrow & \min_W \quad  \text{NLL}(\mathcal{D}; W).
\end{align}

The first term in Eq.~\ref{eq:entropyin} is the average entropy of the true conditional distribution and is therefore independent of $W$.
The last term can be approximated via a Monte-Carlo estimate using the empirical distribution, which results in the negative log likelihood (NLL):
\begin{align}
    -E_{p(X,Y)}\big[\log p(Y\mid W,X)\big] \approx \frac{1}{|\mathcal{D}|}\text{NLL}(\mathcal{D}; W) = -\frac{1}{|\mathcal{D}|} \log p(\mathcal{D} \mid W)
    \label{sm:eq:nll}
\end{align}

Thus minimizing the KL is equivalent to minimizing the NLL, independent of the chosen hypothesis class.

In the following subsections, we indicate the losses derived for each hypothesis class considered in this work. The interested reader can verify that minimizing these losses follows directly from minimizing the NLL under each model assumption.

\subsection{Univariate regression problems}

Here we describe the loss functions used for the 1D toy experiments, as well as the steering angle prediction task, which both have a 1-dimensional outcome: $Y \in \mathbb{R}$.

\paragraph{GLc} This hypothesis class uses a Gaussian likelihood function\footnote{The likelihood function is a function that maps the parameters $W$ to a real value. When we refer to a \textit{Gaussian likelihood}, we mean that this likelihood function is defined by evaluating the dataset $\mathcal{D}$ on input-conditioned Gaussian density functions given by the model $p(Y \mid W, X)$.} and constant variance, the only output of the network is the mean of the Gaussian $\mu(X; W)$, while the variance $\sigma$ is input-independent. In this case the corresponding loss is:
\begin{align}
    \text{NLL}(\mathcal{D}; W) = - \log p(\mathcal{D} \mid W)
    %&\approx - \frac{N}{N_{mb}} \sum_{(x,y) \in \mathcal{B}}\log p(y |W, x)\\
    %&=  - \frac{N}{N_{mb}} \sum_{(x,y) \in \mathcal{B}} \log \mathcal{N} (\mu(x;W), \sigma^2) \\
    %&= - \frac{N}{N_{mb}} \sum_{(x,y) \in \mathcal{B}} \log \Big[ \frac{1}{\sqrt{2 \pi} \sigma} exp \big( - \
    %\frac{(y - \mu(x; W))^2}{2 \sigma^2}\big) \Big] \\
    & \approx \frac{N}{N_{mb}} \sum_{(x,y) \in \mathcal{B}}\frac{(y - \mu(x;W))^2}{2 \sigma^2} + \text{const.} 
\end{align}
where $\mathcal{B}$ denote mini-batches of size $N_{mb}$ and $N$ is the size of the entire dataset. Note, that apart from constant factors, this loss is the commonly used mean-squared error loss.

\paragraph{GL} This hypothesis class uses a Gaussian likelihood function and input-dependent variance, also the variance $\sigma(X;W)$ is outputted by the model. In this case the corresponding loss is:
\begin{align}
    \text{NLL}(\mathcal{D}; W) & \approx  \frac{N}{N_{mb}} \sum_{(x,y) \in \mathcal{B}} \log \sigma(x; W) +
    \frac{(y - \mu(x; W))^2}{2 \sigma(x; W)^2} 
    + \text{const.}
\end{align}

This type of regression model has also been considered, for instance, in \citet{nix:1994:mean:variance, deep:ensembles, kendall:2017:what:uncertainties:do:we:need}. Recently, a method that empirically shows improved calibration in such models has been proposed by \citet{f:Cal:2021:bhatt}.

\paragraph{FL} For models based on normalizing flows, the corresponding loss is:
\begin{align}
    \text{NLL}(\mathcal{D}; W) & \approx  - \frac{N}{N_{mb}} \sum_{(x,y) \in \mathcal{B}} \log p_z(g^{-1} \left(y; f(x,W))\right) + \log \mid \text{det} \, \mathcal{J}_{g^{-1}(y;f(x,W))} \mid \label{eq:nfloss}
\end{align}
where $p_z(Z)$ is the base distribution of the input-dependent flow $g\left(Z;f(X,W) \right)$ with flow parameters $f(X,W)$, and $\mathcal{J}_{g^{-1}(Y;f(X,W))}$ denotes the Jacobian of the inverse of the flow. Note that, in this case, the base network $f(X,W)$ takes the role of a hypernetwork \citep{ha:hypernetworks}.

\paragraph{Bayesian models} To capture approximation uncertainty, we explicitly model a distribution over network parameters using approximate Bayesian inference. In particular, we chose Bayes-by-Backprop \citep{bayes:by:backprop} which employs variational inference for a Gaussian mean-field variational family. The corresponding ELBO \citep{blei:vi:review} objective can be captured in the following loss function:
\begin{align}
     \frac{1}{K} \sum_{k=1}^K \text{NLL}(\mathcal{D}; w_k) + \text{KL}(q(W; \psi) \mid\mid p(W))
\end{align}
where $w_k$ corresponds to weight samples from the approximate posterior $q(W; \psi)$ and $p(W)$ denotes the weight prior.

\subsection{Bivariate regression problems}
In this subsection, we describe the loss functions used for the toy experiments with 2-dimensional outcome: $Y \in \mathbb{R}^2$.

\paragraph{GLc} This hypothesis class uses a Gaussian likelihood function with constant diagonal covariance matrix $\sigma^2 I$. The network outputs two units corresponding to the mean of the Gaussian $\mu(X; W)$ and the corresponding loss function is: 
\begin{align}
    \text{NLL}(\mathcal{D}; W) = - \log p(\mathcal{D} \mid W)
         & \approx \frac{N}{N_{mb}} \sum_{(x,y) \in \mathcal{B}}\frac{||y - \mu(x;W)||_2^2}{2 \sigma^2} + \text{const.} 
\end{align}
where $\mathcal{B}$ denotes mini-batches of size $N_{mb}$ and $N$ is the size of the entire dataset.

\paragraph{GL} This hypothesis class uses a Gaussian likelihood function with input dependent covariance matrix. The network output has five dimensions, two for the mean of the Gaussian $\mu(X; W)$, two for the variances $\sigma_1^2(X;W)$ and $\sigma_2^2(X;W)$ and one for the correlation factor $\rho(X;W)$ between the two components of $Y$. The covariance matrix is given by:
\begin{align}
    \Sigma(x;W) &= \begin{pmatrix}
                        \sigma_1^2(x;W) & \rho\sigma_1\sigma_2(x;W)\\
                        \rho\sigma_1\sigma_2(x;W) & \sigma_2^2(x;W)
                    \end{pmatrix}
\end{align}
and the loss to optimize is as follows:
\begin{align}
    \text{NLL}(\mathcal{D}; W) & \approx  \frac{N}{N_{mb}} \sum_{(x,y) \in \mathcal{B}} \frac{1}{2}\log |\Sigma(x; W)| +
    \frac{1}{2}(y - \mu(x; W))^T\Sigma^{-1}(x; W)(y - \mu(x; W)) 
    + \text{const.}
\end{align}

\paragraph{FL} Models based on 2D normalizing flows rely on the same loss function as the one described in the 1D case (Eq. \ref{eq:nfloss}). 

\section{How aleatoric uncertainty is captured with a Gaussian likelihood function}
\label{sm:sec:glc:aleatoric}

Let us denote the expected value of the ground-truth $p(Y \mid X)$ by $\bar{Y}$ and the variance-covariance matrix by $C_Y$. Note, that the NLL loss is an approximation of the expected KL term: $\text{KL} \big( p(Y \mid X) || p(Y \mid W, X) \big)$, which in turn is known to be moment matching. Thus, for a Gaussian likelihood $p(Y \mid W, X) = \mathcal{N}\big(Y; \mu(X; W), \Sigma(X; W) \big)$, the mean $\mu(X; W)$ and variance-covariance matrix $\Sigma(X; W)$ should at optimality match the $\bar{Y}$ and $C_Y$ of $p(Y \mid X)$.

This can easily be verified by looking at the expression below:

\begin{align}
     & \min_W \mathbb{E}_{p(X)} \Big[ \text{KL} \Big( p(Y \mid X) || \mathcal{N}\big( \mu(X; W), \Sigma(X; W) \big) \Big) \Big] \\
     \Leftrightarrow & \min_W   \mathbb{E}_{p(X) p(Y \mid X)} \bigg[ \frac{1}{2} \log | \Sigma(X; W) | +  \frac{1}{2} \big(Y - \mu(X; W) \big)^T \Sigma^{-1} \big(Y - \mu(X; W) \big) \bigg] \\
     \Leftrightarrow & \min_W   \mathbb{E}_{p(X)} \bigg[ \frac{1}{2} \log | \Sigma(X; W) | +  \frac{1}{2} \big(\bar{Y} - \mu(X; W) \big)^T \Sigma^{-1} \big(\bar{Y} - \mu(X; W) \big)  \nonumber \\
     & \hspace{5em} + \frac{1}{2} \text{tr}\big( \Sigma(X; W)^{-1} C_Y \big) \bigg] 
\end{align}

where the last equality is obtained using the identity for quadratic forms: $\mathbb{E}[\epsilon^T A \epsilon] = \text{tr}(A C_\epsilon) + \bar{\epsilon}^T A \bar{\epsilon}$. The non-negative middle term vanishes if means are matched $\mu(X; W) \equiv \bar{Y}$, leaving the expression

\begin{equation}
    \frac{1}{2} \Big( \log | \Sigma(X; W) | + \text{tr}\big( \Sigma(X; W)^{-1} C_Y \big) \Big) \quad \text{,}
\end{equation}

which, if derived with respect to $\Sigma(X; W)$, becomes

\begin{equation}
    \label{sm:eq:sigma:grad}
    \frac{1}{2} \Big( \Sigma(X; W)^{-1} - \Sigma(X; W)^{-1} C_Y \Sigma(X; W)^{-1} \Big)^T \quad \text{.}
\end{equation}

To obtain this derivative, we made use of the identities $\frac{\partial \log |X|}{\partial X} = (X^{-1})^T$ and $\frac{\partial \text{tr}(A X^{-1} B)}{\partial X} = - (X^{-1} B A X^{-1})^T$. The derivative in Eq.~\ref{sm:eq:sigma:grad} is zero for $\Sigma(X; W) = C_Y$, which, due to the convexity of the KL, corresponds to a minimum.

\newpage 

\section{Quantifying epistemic uncertainty}
\label{sm:sec:uncertainty}

\begin{figure}[h]
    \centering
    \includegraphics[width=.85\textwidth]{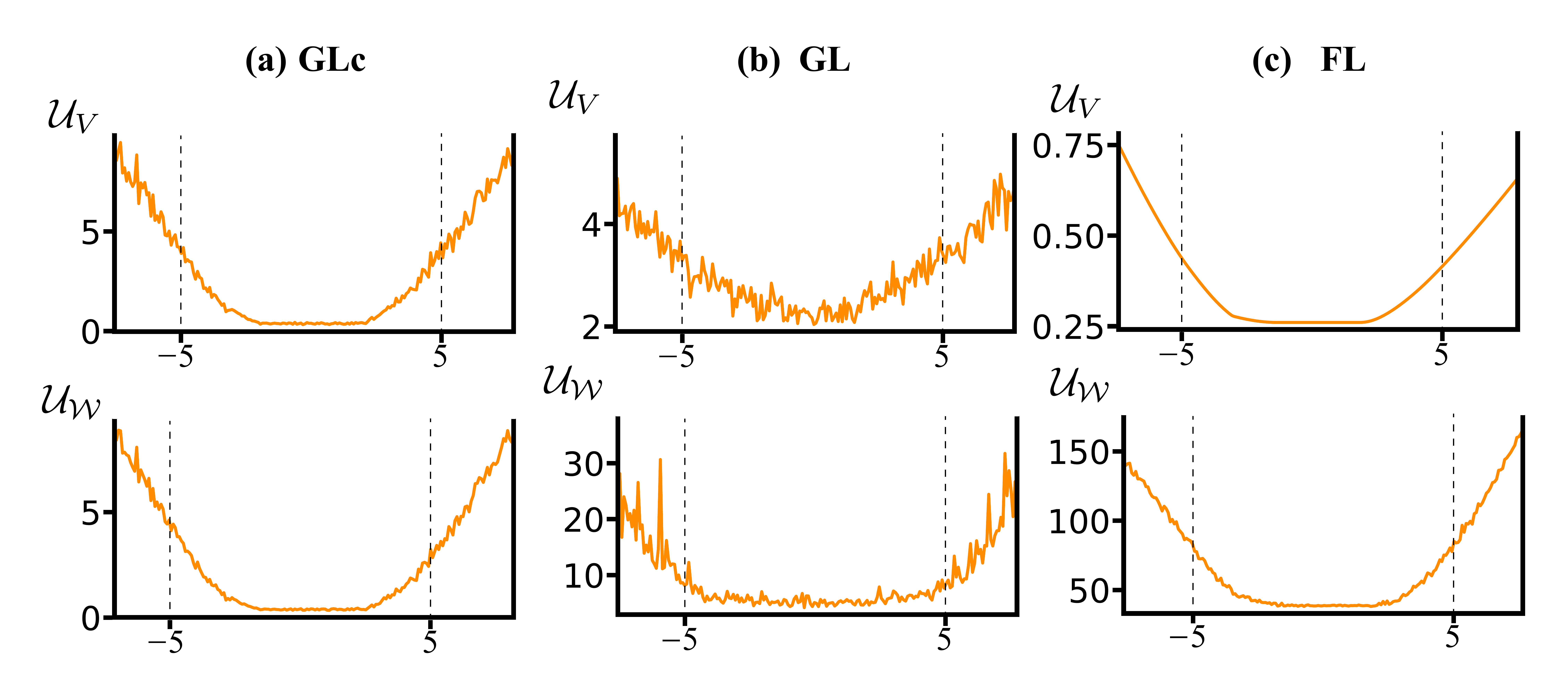}
    \caption{Measures of model disagreement for the 1D mixture of Gaussians experiments under different model classes. \textbf{(a)} Gaussian likelihood with fixed variance, \textbf{(b)} or varying variance, \textbf{(c)} normalizing flow-based likelihood. Interestingly, the average variance in the parameter space $\mathcal{U}_V$, despite not being principled for the normalizing flow as illustrated in cf. SM \ref{sm:sec:uncertainty}, still provides a meaningful estimation similar to $\mathcal{U}_{\mathcal{W}}$.}
    \label{fig:model:disagreement}
\end{figure}

As outlined in Sec.~\ref{sec:intro}, epistemic uncertainty consists of model and approximation uncertainty \citep{hullermeier2021aleatoric:epistemic}. Model uncertainty is in practice often ignored, leaving approximation uncertainty as the sole contributor to the estimated epistemic uncertainty.
In this section, we also make the simplifying assumption that the considered hypothesis space is rich enough to contain a given ground-truth and equate approximation uncertainty with epistemic uncertainty. Epistemic uncertainty is an intrinsic property of the Bayesian formulation,\footnote{But see \citet{owhadi:brittleness:bayesian:inference:2015} for a discussion on the brittleness of such uncertainty estimates.} nevertheless there seems to be no generally accepted agreement on how this type of uncertainty can be quantified in an input-dependent way. The entropy of the posterior predictive distribution is commonly used in the literature as a proxy for epistemic uncertainty:
\begin{equation}
    \mathcal{H}[p(Y \mid \mathcal{D},X)] = \int_y p(y \mid \mathcal{D},X) \log p(y \mid \mathcal{D},X) dy \, .
    \label{eqn:entropy_predictive}
\end{equation} 
However, it does not allow for a disentanglement of epistemic and aleatoric uncertainty and should therefore not be used as a measure of the former. 
To overcome this limitation, we study different possible measures of epistemic uncertainty based on the intuition that a summary statistic of $p(Y \mid \mathcal{D},X)$ representative of the epistemic uncertainty should capture the diversity between the individual predictive distributions $p(Y \mid W, X)$ sampled from the parameter posterior. Whenever the mapping between parameters $\theta = f(X;W) \in \mathbb{R}^m$ and hypothesis $g: \theta \mapsto p(Y|W,X)$ is injective (for instance in GL and GLc) the aforementioned diversity can be easily measured as the average variance across the dimension of the parameters: 
\begin{equation}
    \mathcal{U}_V(x) = \frac{1}{m} \sum_i \text{Var}_{p(W|\mathcal{D})}[\theta_i] \, .
\end{equation}
However this might become ambiguous for the normalizing flow case where $g$ is not identifiable and the map is not unique anymore. Under these circumstances, we propose to use a divergence based measure of disagreement defined as: 
\begin{equation}
    \mathcal{U}_D(x) = \mathbb{E}_{p(W|\mathcal{D})} \left[ D(p(Y \mid W, X) \mid \mid p(Y \mid \mathcal{D}, X)) \right] \, , \end{equation}
where $D(\cdot \mid \mid \cdot)$ is a statistical divergence. In our analysis we consider the Wasserstein distance $\mathcal{U}_{\mathcal{W}}$. The results for the different measures of model disagreement are reported in Fig. \ref{fig:model:disagreement}. 

For all models, one can see that model disagreement is lowest in-distribution, independent on the considered measure. It is important to notice, that $\mathcal{U}_V(x)$ quantities cannot be compared between models as the parameter spaces $\theta$ are different.

\section{Out-of-distribution detection in the context of regression}
\label{sm:sec:ood}

Out-of-distribution (OOD) detection \citep{zimek2018outlier:detection} is concerned with the question of whether an input stems from the input distribution $p(X)$. The most natural approach to tackle this question is to model $p(X)$ directly, which is challenging for high-dimensional input data \citep{nalisnick:2018ood:generative}.
By contrast, uncertainty-based OOD detection uses summary statistics of $p(Y \mid \mathcal{D}, X)$ to determine whether an input is considered OOD. Some works perform this task by considering only deterministic networks \citep{ood:baseline}, where $p(Y \mid \mathcal{D}, X)$ is approximated using, for instance, the MAP estimate. These approaches have thus no access to epistemic uncertainty estimates. Therefore, and unless explicitly trained on OOD data, they rely on uncalibrated uncertainty estimates, as common loss functions (e.g., Eq.~\ref{sm:eq:nll}) only calibrate inside the support of $p(X)$. Epistemic uncertainty can be heuristically introduced by using an ensemble of networks \citep{deep:ensembles}, or in a more principled manner by applying approximate Bayesian inference \citep{louizos:multiplicative:nf:2017}. However, whether epistemic uncertainty is useful for the task of OOD detection severely depends on the employed function space prior as discussed in \citet{henning2021bayesian}. Simply speaking, the summary statistics used to decide whether an input is OOD should reflect properties intrinsic to $p(X)$. Under what circumstances such link can be established is not well studied \citep{henning2021bayesian}, which means that uncertainty-based OOD methods can in practice only be empirically tested for their reliability on arbitrarily chosen OOD datasets.

Under the clear premise that uncertainty-based OOD detection is likely not principled for the methods considered in this study, we would like to discuss observed differences in OOD uncertainty quantification. As GLc has no means to exhibit input-dependent uncertainty, it cannot be used for OOD detection. This is not the case for GL and FL models, but as these models are trained with the NLL (cf. SM \ref{sm:sec:loss}), uncertainty is only calibrated in-distribution. Therefore, we have no explicit control over OOD uncertainties, as becomes apparent in Fig. \ref{fig:bimodal:1D:experiments}c, where OOD uncertainty is lowest.

More interesting are the cases where some epistemic uncertainty is explicitly captured as parameter uncertainty by using Bayesian neural networks (BNN). For the case of a Gaussian likelihood such as GLc in combination with a Gaussian process prior using an RBF kernel, \citet{henning2021bayesian} showed that the variance of the predictive posterior can be related to $p(X)$. Interestingly, the formula for this variance only depends on the likelihood parameters and training inputs, but not the training targets, which means that model misspecification as in Fig. \ref{fig:bimodal:1D:experiments}a would not harm the result that epistemic uncertainty is reflective of $p(X)$. In general, however, it is to be expected that the induced function space prior, the choice of likelihood and the training data (incl. target values) together will shape the behavior of epistemic uncertainty in OOD regions. Therefore, unless future work provides a theoretical justification, uncertainty-based OOD detection should be considered with care also for regression tasks.

\section{Experimental details}
\label{sm:sec:experiments}

We use the publicly available code base from \citet{posterior:replay:2021:henning:cervera}, which provides a PyTorch-based interface for working with hypernetworks.\footnote{\url{https://github.com/chrhenning/hypnettorch}} This interface allows to easily parametrize flow-based models. We use neural spline flows \citep{neural:spline:flows} for all FL experiments, using the implementation from \citet{nflows}.

\subsection{1D experiments}
\label{sm:exp:oned}

Our 1D experiments consist of a dataset with $X, Y \in \mathbb{R}$, where $Y$ is sampled from one of the two conditional distributions $p(Y \mid X)$ described below.

In the simple \textbf{unimodal} setting introduced by \cite{hernandez:lobato}, $y = x^3 + \epsilon$, where $\epsilon \sim \mathcal{N} (0, \sigma^2)$ and $\sigma^2=9$. The ground-truth likelihood is therefore $p(y \mid x) = \mathcal{N} (x^3, 9)$. The training interval was defined to be $[-4, 4]$, and test points are equidistantly placed in the same interval. For plotting, however, we evaluate the model on an interval $[-6, 6]$, such that behavior on out-of-distribution data could be studied. The dataset consists of 20 training points and 50 test points.

In the \textbf{bimodal} setting introduced in this paper,
the ground-truth likelihood is given as $p(y \mid x) = \frac{1}{2}  \big( \mathcal{N} (x^3 - 50, 9) + \mathcal{N} (x^3 + 50, 9)\big)$. The training and test intervals are identical to the unimodal dataset. In this case, the plots were obtained with 1000 training points unless noted otherwise.

\subsection{2D experiments}

To highlight the limitations of standard MSE training for capturing correlations between output dimensions, we consider a regression problem with 2D output space.

\label{sm:exp:twod}
\subsubsection{Dataset}
The bivariate experiments use a dataset $X,Y$ where $X \in \mathbb{R}$ and $Y \in \mathbb{R}^2$. A particular \textbf{bimodal} conditional distribution $p(Y\mid X)$ is studied. Samples are obtained by sampling the bimodal distribution $\frac{1}{2}  \big( \mathcal{N} (X^3 - 15, \Sigma) + \mathcal{N} (X^3 + 15, \Sigma)\big)$ where $\Sigma = \text{diag}([300,20])$, samples are then rotated in the $(Y1,Y2)$-plane around the mean $(X^3, X^3)$ by the angle $\pi/4$. The obtained conditional distribution $p(Y \mid X)$ is still bimodal but both the marginals $p(Y_1 \mid X)$ and $p(Y_2 \mid X)$ look Gaussian. In addition, the total distribution has a diagonal variance with the same variance in both directions whereas in both modes the components of $y$ are strongly correlated (Fig. \ref{fig:bimodal:2D:experiments}).

\subsubsection{Bivariate normalizing flows}
2D normalizing flows rely on the succession of coupling transform layers, each composed of both a transformer and a conditioner to transform a 2D base distribution $p(U)$ into a 2D target distribution $p(Y)$. 
Similarly to the 1D case, the transformer is a univariate diffeomorphism which transforms one of the components of an input distribution, while the other component is kept unchanged.
To introduce dependency between the two components, the conditioner takes the unchanged component as input and outputs the weights of the transformer.
In order to make the target distribution $p(Y)$ conditional on $X$, the parameters of each layer (the parameters of the conditioner) are obtained from a hypernetwork $h(X;W)$ that takes $X$ as input. The target distribution $p(Y)$ therefore turns into a conditional $p(Y \mid X)$.

\subsubsection{Bivariate regression}

\begin{figure}[h]
    \centering
    \includegraphics[width=0.85\textwidth]{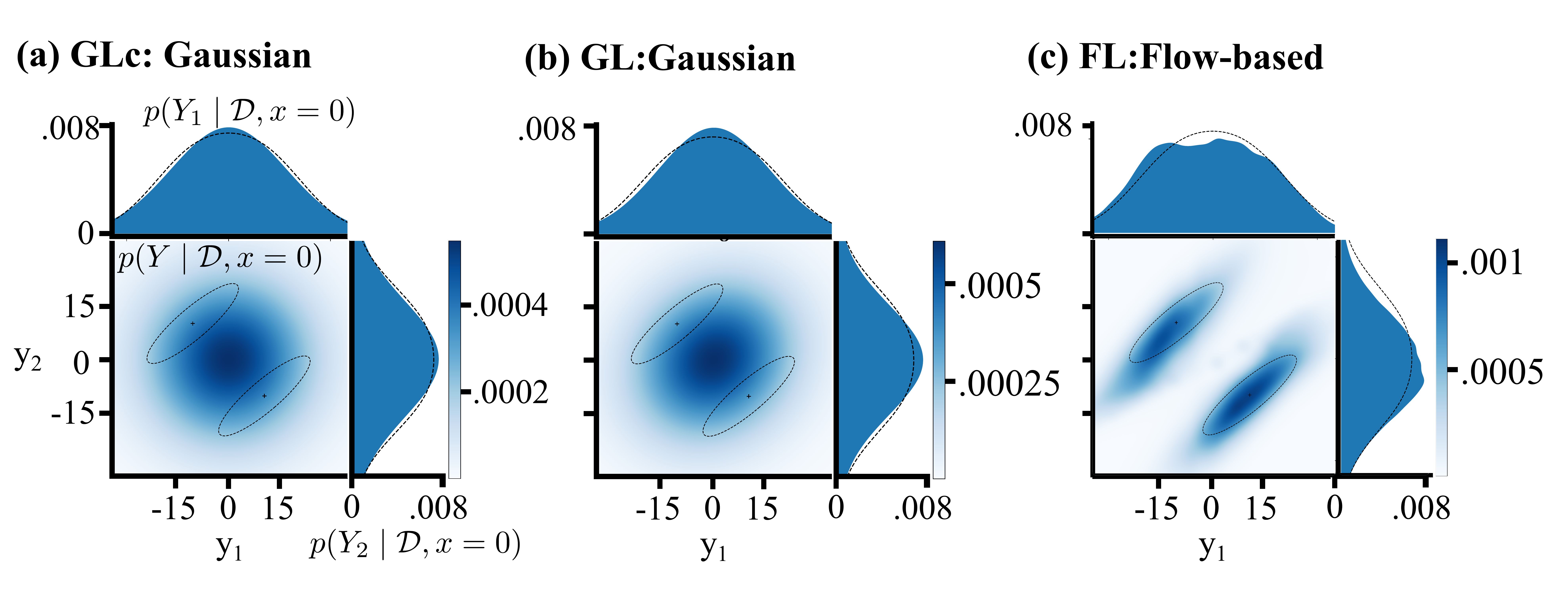}
    \caption{Bivariate regression on a bimodal conditional distribution. Center plots correspond to the bivariate conditional distribution at an example value of $x=0$: $p(Y \mid \mathcal{D},x=0)$. The upper and right plots correspond to the associated 1D marginal distributions. In the center plots, '+' markers indicate the mean of each mode of the true distribution, and the dashed ellipses indicate the standard deviation contours of each mode. On the marginal plots, a dashed line indicates the true marginal distribution. The blue areas indicate the estimated conditional distribution $p(Y \mid \mathcal{D},x=0)$.  \textbf{(a)} Gaussian likelihood with fixed covariance, \textbf{(b)} or varying covariance, \textbf{(c)} normalizing flow-based likelihood.}
    \label{fig:bimodal:2D:experiments}
\end{figure}

The results obtained for all different modeling choices on the bivariate regression problem can be found in Fig. \ref{fig:bimodal:2D:experiments} and Table \ref{tab:bimodal:2D:experiments}.

Using a Gaussian likelihood leads to matching at best the first two moments of the conditional distribution at each point $x$, better visualized on the marginals (Fig. \ref{fig:bimodal:2D:experiments}a and b). In doing so, the model fails to capture the strong correlation that exists within each mode (Fig. \ref{fig:bimodal:2D:experiments}a and b). Using models that induce a more flexible likelihood is necessary in such cases, and indeed lead to lower NLL (Table \ref{tab:bimodal:2D:experiments}).

\begingroup
\setlength{\tabcolsep}{2.5pt}
\begin{table*}[h]
 \centering
  \caption{Test NLL and MSE of 2D bimodal experiments (Mean $\pm$ SEM, $n=10$).}
  \vskip 0.15in
  \begin{small}
  \begin{tabular}{lcc} \toprule
    & Test NLL & Test MSE \\ 
    \midrule\midrule
    \textbf{GLc} & $8.46 \pm 0.04$ & $550.13 \pm 25.99$\\
    \textbf{GL}  & $8.45 \pm 0.04$ & N/A \\
    \textbf{FL}  & $7.97 \pm 0.05$ & N/A \\
    \bottomrule
  \end{tabular}
  \label{tab:bimodal:2D:experiments}
  \end{small}
\end{table*}
\endgroup

\subsection{Steering angle prediction experiments}
\label{sm:exp:udacity}

For the steering angle prediction task, we use the dataset provided with the Udacity self-driving car challenge 2.\footnote{The dataset can be downloaded here \url{https://github.com/udacity/self-driving-car/tree/master/datasets/CH2}.} We only use the front camera images and steering angles from the official training set for training our models. We input plain RGB images resized to 256 x 192 into the model, without applying any data augmentation. As base network $f(X; W)$ we use a Resnet-18 \citep{he2016resnet}, without any pre-training. The results obtained for the different modeling choices can be found in Table \ref{tab:steering}.

\begingroup
\setlength{\tabcolsep}{2.5pt}
\begin{table*}[h]
 \centering
  \caption{Test NLL and MSE of steering angle prediction experiments (Mean $\pm$ SEM, $n=10$).}
  % These results were obtained by searching for 30 epochs and looking at final validation loss.
  \vskip 0.15in
  \begin{small}
  \begin{tabular}{lcc} \toprule
    & Test NLL & Test MSE \\ 
    \midrule\midrule
    \textbf{GLc} & -2.00 $\pm$ 0.05 & 0.00079 $\pm$ 0.00004 \\
    \textbf{GL}  & -2.76 $\pm$ 0.09 & N/A \\
    \textbf{FL}  & -2.91 +- 0.10 & N/A \\
    \bottomrule
  \end{tabular}
  \label{tab:steering}
  \end{small}
\end{table*}
\endgroup

As expected, GLc achieves the highest loss on a held-out validation set in this task (and thus, the worst calibration). Extending the model to also generate input-dependent variance leads to a noticeable reduction in the loss in GL, which can be further reduced by considering more flexible conditionals with the FL model. 

\section{Additional Experiments}
\label{sm:sec:add:exp}

\subsection{1D unimodal Gaussian experiments}

\begin{figure}[h!]
    \centering
    \includegraphics[width=0.85\textwidth]{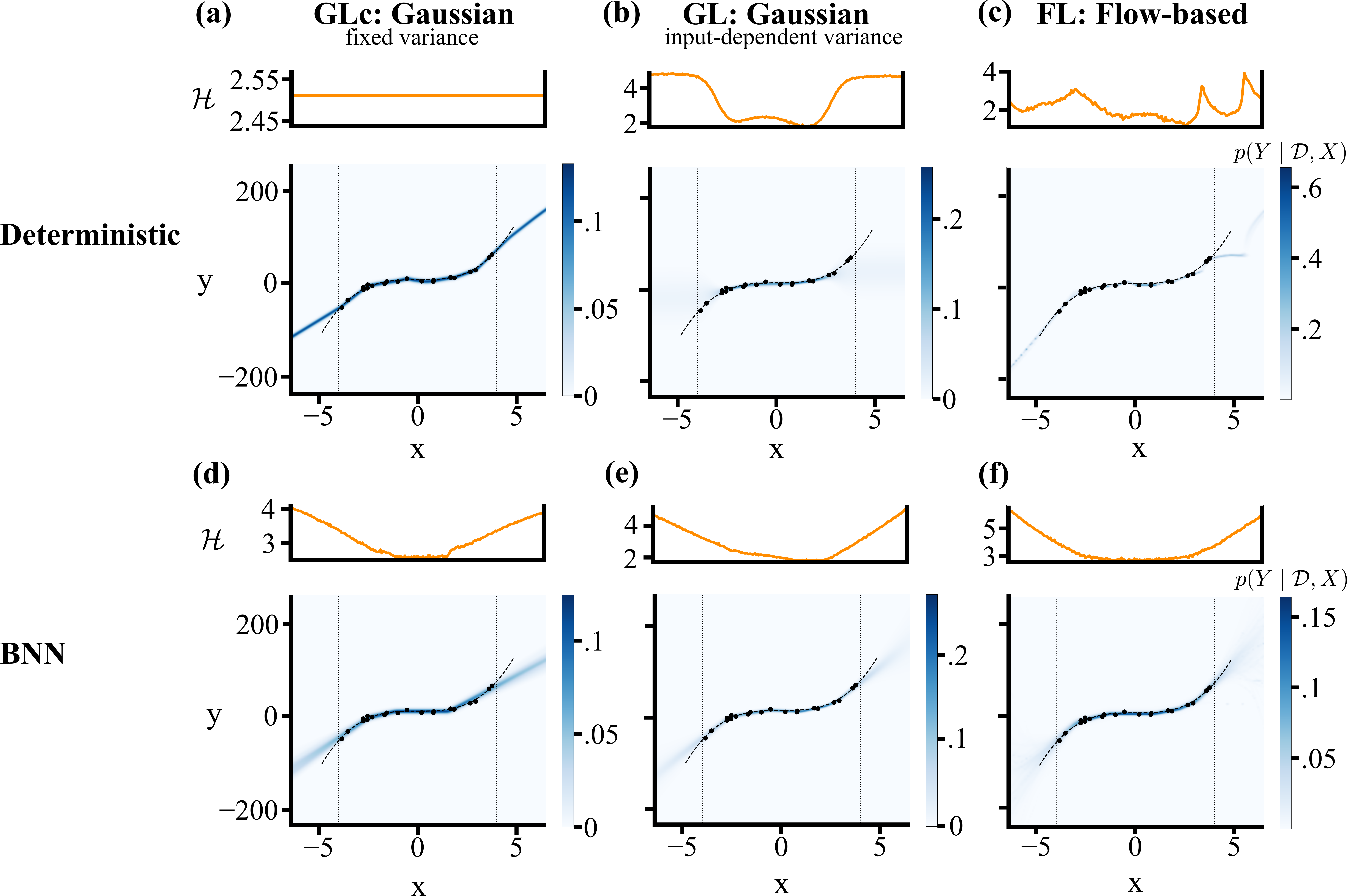}
    \caption{Differential entropy $\mathcal{H}$ and predictive distributions for the 1D unimodal Gaussian experiments under different model classes. \textbf{(a)} Gaussian likelihood with fixed variance, \textbf{(b)} or varying variance, \textbf{(c)} normalizing flow-based likelihood. \textbf{(d), (e)} and \textbf{(f)} Same as above but using BNNs instead. The dashed line indicates the mean of the ground-truth, and the vertical dotted lines indicate the boundaries of the in-distribution range. Dots indicate the 20 training points.}
    \label{fig:unimodal:1D:experiments}
\end{figure}

In Fig. \ref{fig:unimodal:1D:experiments} and Table \ref{sm:tab:1D:unimodal} we report results for the classic unimodal 1D regression task introduced in \cite{hernandez:lobato}, using 20 training points as commonly done.
Due to the simplicity of this dataset, even the simplest of models (GLc) can properly capture the ground-truth in-distribution. 
As commonly observed \citep[e.g., ][]{louizos:multiplicative:nf:2017}, GLc in a Bayesian setting exhibits high diff. entropy on OOD data, creating the impression that BNNs are useful OOD detectors (cf. SM \ref{sm:sec:ood}). This behavior is also present in the deterministic setting for GL, which can capture input-dependent uncertainty. Interestingly, the flexible FL models matches in-distribution data very well but exhibits arbitrary behavior outside the training range.

\begingroup
\setlength{\tabcolsep}{2.5pt}
\begin{table*}[t]
 \centering
  \caption{Test NLL (normalized by the number of samples) and MSE of 1D unimodal Gaussian experiments (Mean $\pm$ SEM, $n=10$).}
  \vskip 0.15in
  \begin{small}
  \begin{tabular}{lcc} \toprule
    & Test NLL & Test MSE \\ 
    \midrule\midrule
    \textbf{GLc} & 2.70 $\pm$  0.01  &  12.22 $\pm$  0.10  \\
    \textbf{GL}  & 3.36 $\pm$  0.03 & N/A \\
    \textbf{FL}  & 3.43 $\pm$  0.34  & N/A \\
    \midrule
    \textbf{BNN + GLc} & 3.67 $\pm$  0.03 & N/A \\ %29.68 $\pm$  0.50 \\
    \textbf{BNN + GL}  & 3.66 $\pm$  0.03 & N/A \\
    \textbf{BNN + FL}  & 5.46 $\pm$  0.59  & N/A \\
    \bottomrule
  \end{tabular}
  \label{sm:tab:1D:unimodal}
  \end{small}
\end{table*}
\endgroup

When looking at the quantitative results obtained, it is interesting to note that the simplest model, GLc, achieves lowest loss in this experiment, highlighting the importance of choosing a simple model whenever possible. Similarly, the Bayesian extension of the models does not provide any performance benefits in this simple dataset, showcasing the increased costs of training more complex models.

\subsection{1D mixture of Gaussian experiments}
\begin{figure}[t]
    \centering
    \includegraphics[width=0.85\textwidth]{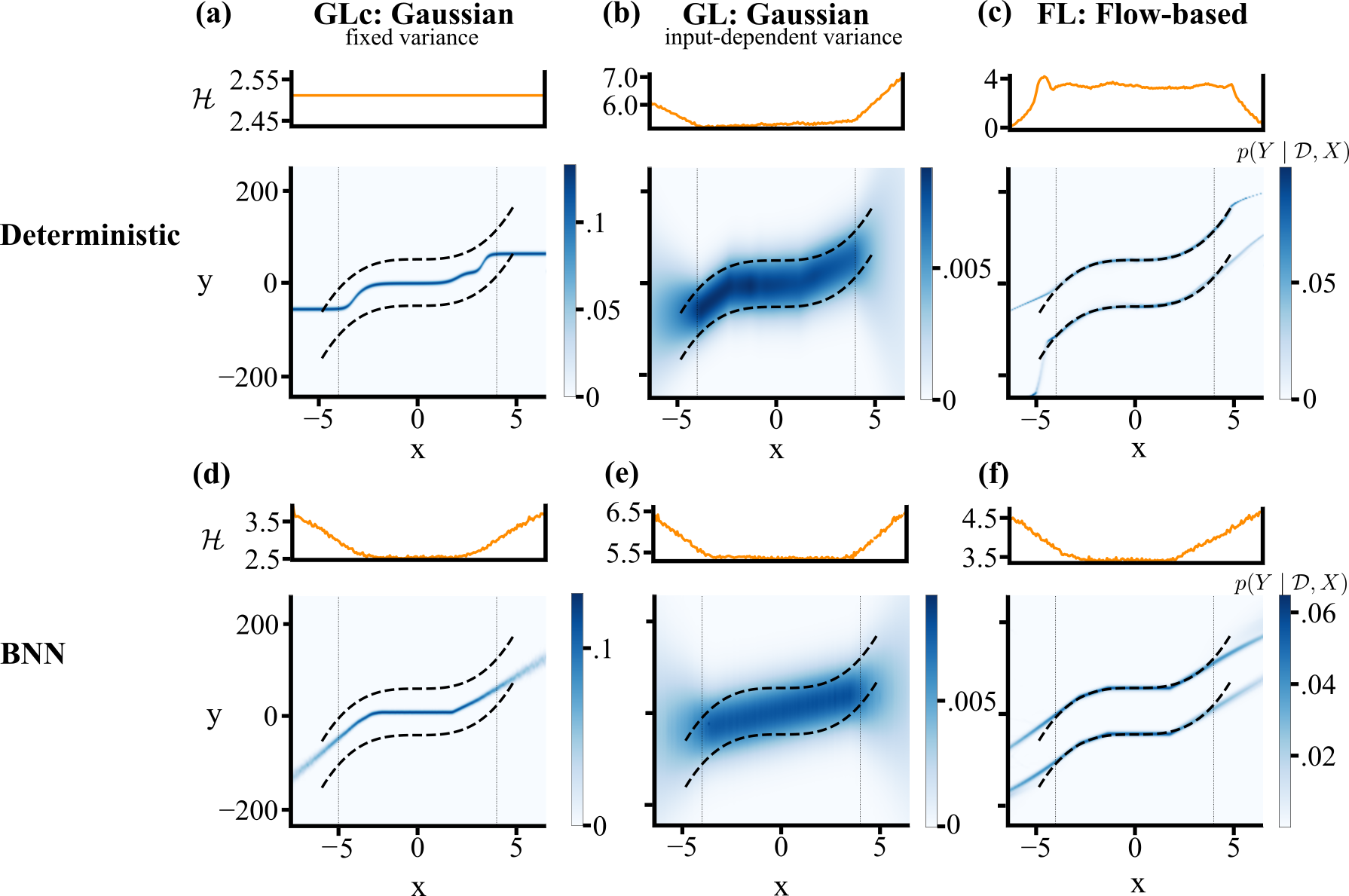}
    \caption{Differential entropy $\mathcal{H}$ and predictive distributions for the 1D mixture of Gaussian experiments with 1000 training points under different model classes. Same setting as in Fig. \ref{fig:bimodal:1D:experiments}. \textbf{(a)} Gaussian likelihood with fixed variance, \textbf{(b)} or varying variance, \textbf{(c)} normalizing flow-based likelihood. \textbf{(d), (e)} and \textbf{(f)} Same as above but using BNNs instead. Dashed lines indicate the means of the two modes in the ground-truth, and the vertical dotted lines indicate the boundaries of the in-distribution range.}
    \label{fig:bimodal:1D:1000pts:experiments}
\end{figure}

\begin{figure}[t]
    \centering
    \includegraphics[width=0.85\textwidth]{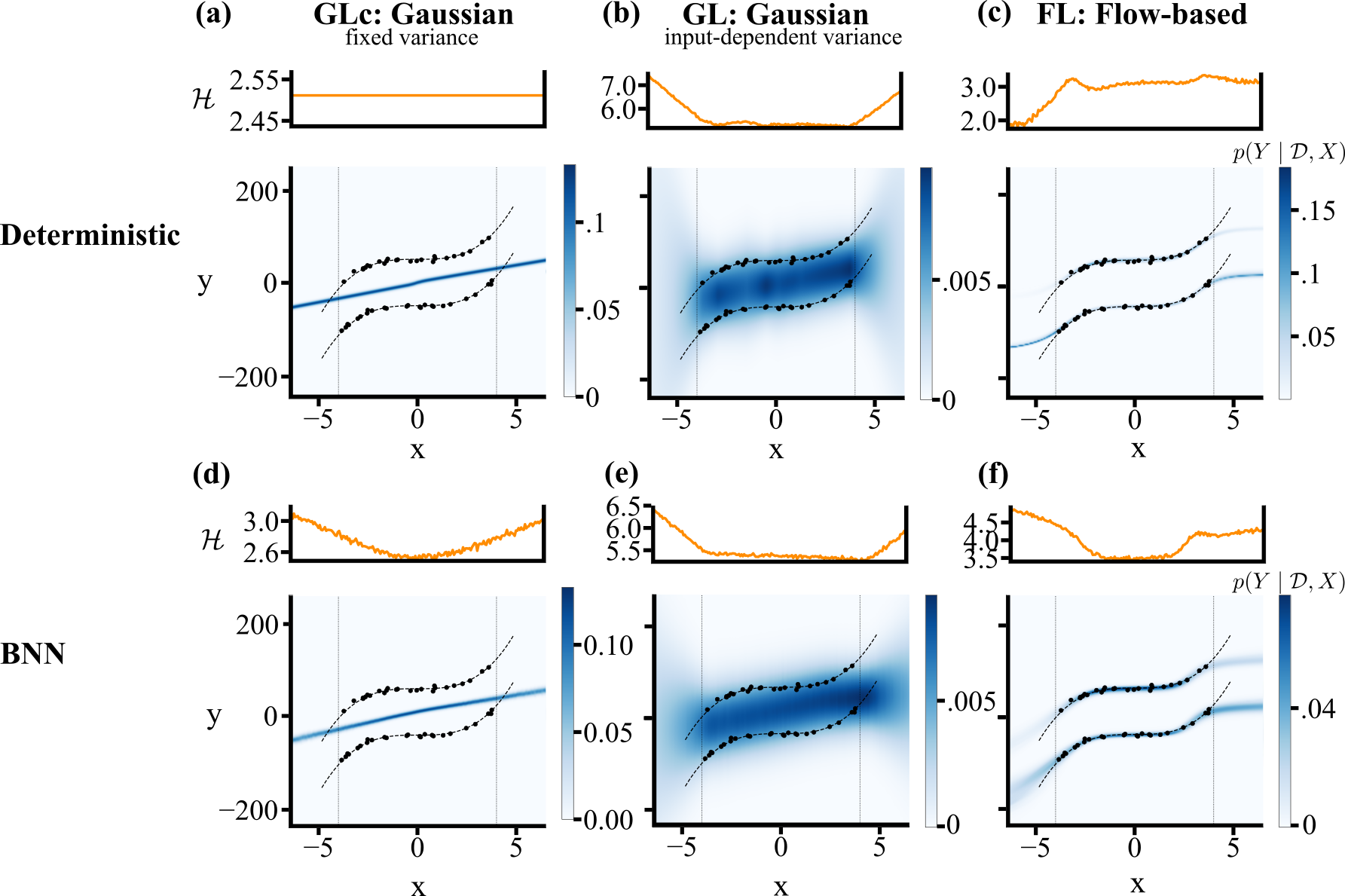}
    \caption{Differential entropy $\mathcal{H}$ and predictive distributions for the 1D mixture of Gaussian experiments under different model classes. Same setting as in Fig. \ref{fig:bimodal:1D:experiments} but with 50 training points instead of 1000. \textbf{(a)} Gaussian likelihood with fixed variance, \textbf{(b)} or varying variance, \textbf{(c)} normalizing flow-based likelihood. \textbf{(d), (e)} and \textbf{(f)} Same as above but using BNNs instead. Dashed lines indicate the means of the two modes in the ground-truth, and the vertical dotted lines indicate the boundaries of the in-distribution range. Dots indicate the 50 training points.}
    \label{fig:bimodal:1D:20pts:experiments}
\end{figure}

In this section, we report complementary results for the 1-dimensional mixture of Gaussian experiments. Compared to Fig. \ref{fig:bimodal:1D:experiments}, in Fig. \ref{fig:bimodal:1D:1000pts:experiments} we show the Bayesian predictive posterior also for the GL and FL models. Additionally, to better understand the impact of the number of training instances we repeated the previous experiment using 50 training data points instead of 1000. The results are shown in Fig. \ref{fig:bimodal:1D:20pts:experiments} and a visual inspection of them suggests that the normalizing flow model can still capture the ground truth distribution. GL also seems capable to capture the mean and variance quite well, while GLc struggles to correctly capture the mean. 

Quantitative results can be observed in Table \ref{sm:tab:1D:mixture}. This example clearly illustrates the failure of simple models like GLc to capture the ground-truth (as measured by the NLL). In these more complex problems, a flow-based model is without question the best choice and leads to the best results. Interestingly, the Bayesian extension only provides a marginal improvement (as shown by the difference between FL and BNN+FL).

\begingroup
\setlength{\tabcolsep}{2.5pt}
\begin{table*}[t]
 \centering
  \caption{Test NLL and MSE of 1D mixture of Gaussians experiments with 50 training points (Mean $\pm$ SEM, $n=10$).}
  \vskip 0.15in
  \begin{small}
  \begin{tabular}{lcc} \toprule
    & Test NLL & Test MSE \\ 
    \midrule\midrule
    \textbf{GLc} & 144.46 $\pm$  0.04 & 2564.01 $\pm$ 0.64 \\ % 2564.01 $\pm$  0.64  \\
    \textbf{GL}  & 5.40 $\pm$  0.01  & N/A \\
    \textbf{FL}  & 3.82 $\pm$  0.40 & N/A \\
    \midrule
    \textbf{BNN + GLc} & 144.86 $\pm$  0.04 & N/A \\ % 2571.17 $\pm$  0.69  \\
    \textbf{BNN + GL}  & 5.41 $\pm$  0.01  & N/A \\
    \textbf{BNN + FL}  & 3.75 $\pm$  0.35 & N/A \\
    \bottomrule
  \end{tabular}
  \label{sm:tab:1D:mixture}
  \end{small}
\end{table*}
\endgroup

\newpage

\begin{figure}[t!]
    \centering
    \includegraphics[width=0.26\textwidth]{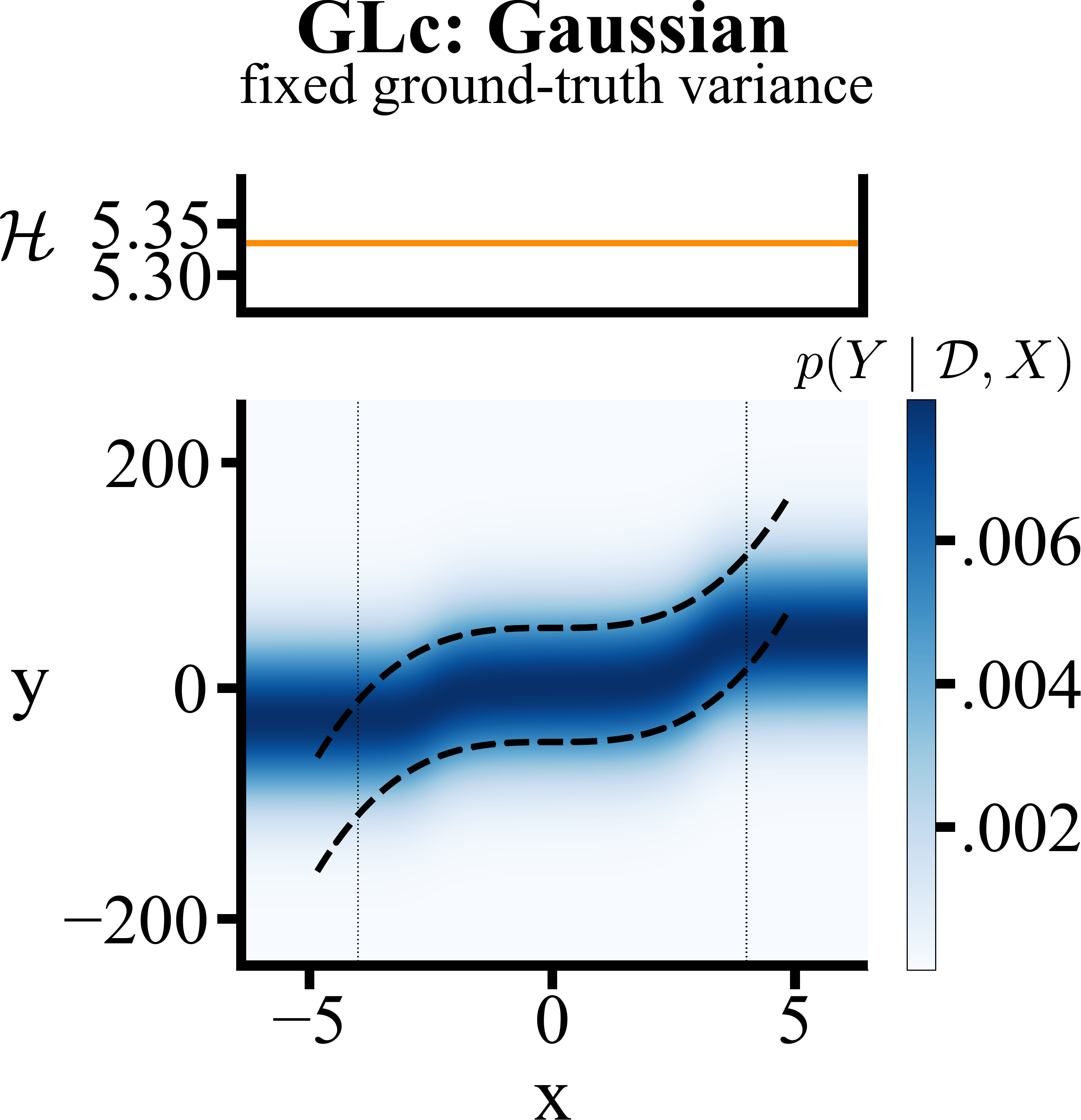}
    \caption{Differential entropy $\mathcal{H}$ and predictive distributions for the 1D mixture of Gaussian experiments for a Gaussian likelihood function with fixed variance that matches the variance of the ground-truth $p(Y \mid X)$. Dashed lines indicate the means of the two modes in the ground-truth, and the vertical dotted lines indicate the boundaries of the in-distribution range.}
    \label{fig:bimodal:1D:true:std}
\end{figure}

We also explored the behavior of GLc for the bimodal 1D toy dataset where, instead of arbitrarily setting the variance, it is chosen to match that of the ground-truth (Fig. \ref{fig:bimodal:1D:true:std}). The variance of the mixture model is given by:
\begin{equation}
    \sigma_\text{MM}^2 = \frac{1}{2}(\sigma_A^2 + \sigma_B^2 ) + \frac{1}{4}(\mu_A(x) - \mu_B(x))^2 = 3^2 + \frac{1}{4}100^2 = 2509
\end{equation}
where $A$ and $B$ denote the two modes of the mixture. As expected, GLc still captures the ground-truth mean well in-distribution.
As opposed to the results in Fig. \ref{fig:bimodal:1D:1000pts:experiments}, this example highlights the importance of properly setting the variance in GLc, to avoid creating false impressions of certainty in the predictions.
% the entropy can be analytically computed in this case as $\frac{1}{2} \log(2 \pi \sigma_{MM}^2) + \frac{1}{2}$.

\end{document}